\documentclass[10pt,twocolumn,letterpaper]{article}

\usepackage{cvpr}              

\usepackage{graphicx}
\usepackage{booktabs}
\usepackage{amsthm,amsmath,amssymb}
\usepackage{algorithm}
\usepackage{algorithmic}
\usepackage{enumitem}
\usepackage{multirow}
\usepackage{bbm}
\usepackage{color}
\usepackage{subcaption}
\usepackage{amsmath}
\usepackage{amssymb}
\usepackage{booktabs}
\usepackage{pifont}
\usepackage{xcolor}
\usepackage{color}
\usepackage{soul}
\usepackage[accsupp]{axessibility}
\usepackage[pagebackref,breaklinks,colorlinks]{hyperref}

\usepackage[capitalize]{cleveref}
\crefname{section}{Sec.}{Secs.}
\Crefname{section}{Section}{Sections}
\Crefname{table}{Table}{Tables}
\crefname{table}{Tab.}{Tabs.}

\def\cvprPaperID{4710} 
\def\confName{CVPR}
\def\confYear{2022}

\begin{document}

\title{Exploring Anchor-based Detection for Ego4D Natural Language Query}

\author{Sipeng Zheng$^1$\\
Renmin University of China\\
{\tt\small zhengsipeng@ruc.edu.cn}
\and
Qi Zhang$^1$\\
Renmin University of China\\
{\tt\small zhangqi1996@ruc.edu.cn}
\and
Bei Liu$^2$\\
Microsoft Research\\
{\tt\small bei.liu@microsoft.com}
\and
Qin Jin
\thanks{corresponding author}\\
Renmin University of China\\
{\tt\small qjin@ruc.edu.cn}
\and
Jianlong Fu
\thanks{corresponding author}\\
Microsoft Research\\
{\tt\small jianf@microsoft.com}
}
\maketitle

\begin{abstract} 
In this paper we provide the technique report of Ego4D natural language query challenge in CVPR 2022.
Natural language query task is challenging due to the requirement of comprehensive understanding of video contents.
Most previous works address this task based on third-person view datasets while few research interest has been placed in the ego-centric view by far.
Great progress has been made though, we notice that previous works can not adapt well to ego-centric view datasets e.g., Ego4D mainly because of two reasons: 
1) most queries in Ego4D have a excessively small temporal duration (e.g., less than 5 seconds); 
2) queries in Ego4D are faced with much more complex video understanding of long-term temporal orders.
Considering these, we propose our solution of this challenge to solve the above issues.

\end{abstract}

\section{Task Introduction}
Natural Language Query (NLQ) \cite{anne2017localizing} has drawn increasing interest due to its essential role for video understanding and serves as a stepping stone for numerous tasks including video-text retrieval \cite{chen2020fine}, video summarization \cite{song2015tvsum,chu2015video} and temporal action localization \cite{zhao2017temporal}.
Given a text query, this task aims to localize both starting and ending time of a segment in an untrimmed video.
A crucial distinction between NLQ and conventional video grounding tasks such as object detection \cite{ren2015faster} or relation detection \cite{zhang2017visual} is that it not only requires accurate recognition of objects, scenes and action, but also comprehension of spatio-temporal relationship between human and objects.

To address this task, the dominant NLQ paradigm can be viewed as a two-stage pipeline: first it computes video and text query embeddings using off-the-shelf models like C3D and Glove, then an inter-modal interaction module will be applied to fuse the pre-computed video and text embeddings. Finally, the timestamps of query are predicted based on the fused embeddings. Such paradigm has achieved great progress in traditional benchmark datasets including ActivityNet, Charades-STA and TACoS. 
However, it achieves poor performance on Ego4D dataset on the opposite.
We suggest the main reasons are two-folds:
\textbf{1) First}, the temporal duration of queries in Ego4D is much more shorter than previous benchmark datasets. According to our statistics, more than $50\%$ queries are less than 5 seconds meanwhile most untrimmed videos are longer than 8 minutes.
Therefore, even a tiny error will lead to a fail prediction. 
\textbf{2) Second}, queries in Ego4D dataset requires long-term temporal orders. (e.g., ``Where was the white glue bottle before I picked it up'').

In the following section, we give a brief description of our method to solve these challenges.

\begin{figure}[t]
	\begin{center}
		\includegraphics[width=1\linewidth]{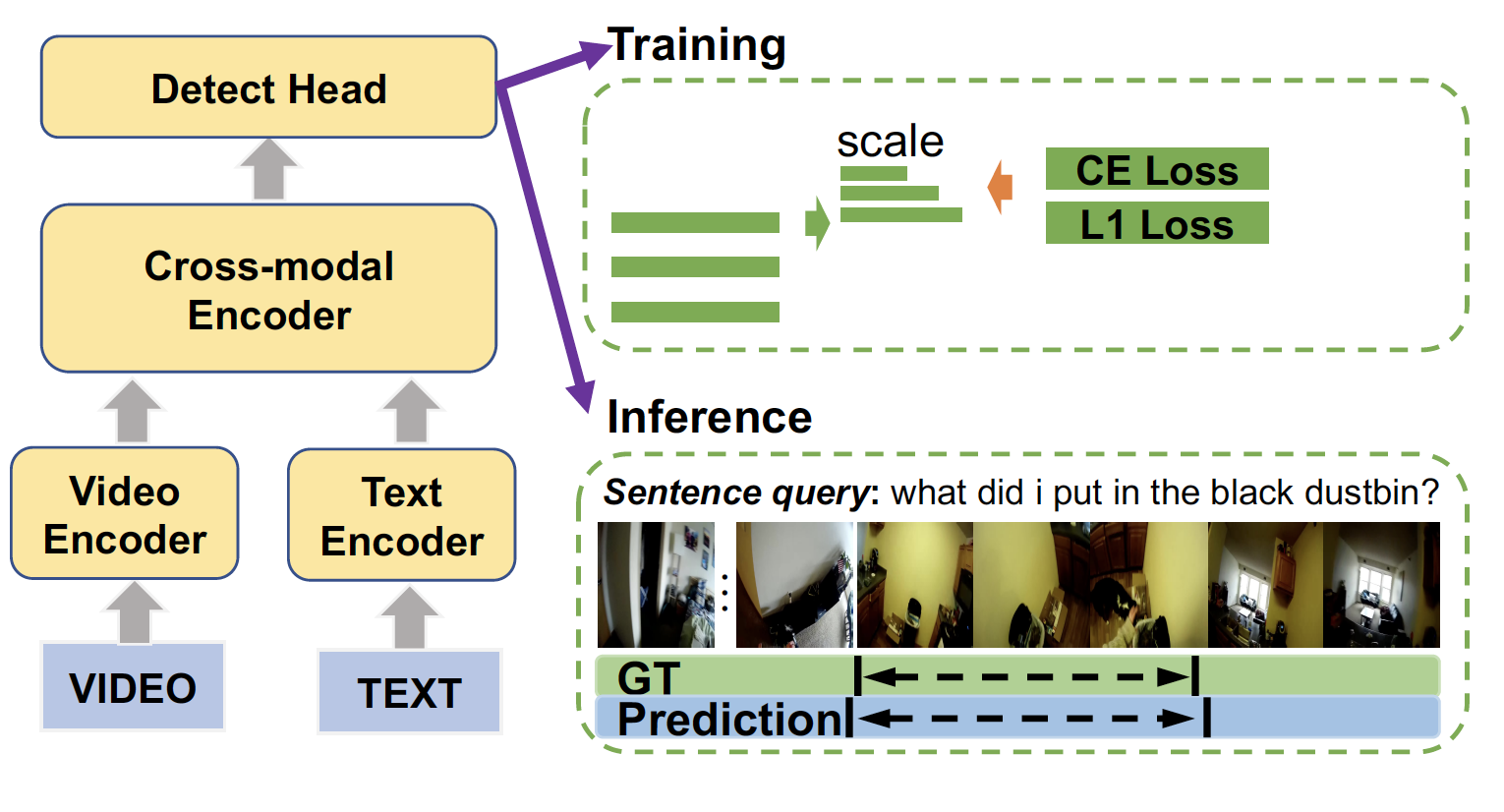}
	\end{center}
	\vspace{-8pt}
	\caption{Overview of our framework  and an example of Ego4D query.}
	\label{fig:case}
\end{figure}
\section{Methodology}
The natural language query (NLQ) aims to localize the matching temporary segment in an untrimmed video given a query sentence.
The video with $\hat{T}$ consecutive frames is denoted as $V=\{I_t\}_{t=1}^{\hat{T}}$ and the segment to be grounded is denoted as $\{I_t\}_{t=t^s}^{t^e}$, where $t^s$ and $t^e$ are the starting and ending timestamps of the segment.
We further parse the query sentence $Q$ into $L$ word tokens as $Q=\{e_i\}_{i=1}^{L}$.

\subsection{Overall Framework}
\label{sec:vl_encode}
Our method can be viewed as a two-stage framework, which can be seen in Figure~\ref{fig:case}.

In the first stage, we sample $T$ key frames from the total $\hat{T}$ frames of untrimmed video, then we extract $T$ frame-level visual features using off-the-shelf visual backbones. 
Since an Ego4D untrimmed video is generally longer than 8 minutes while most queries are within 10 seconds, we sample as many frames as possible to enable our model adapt to longer videos.
Meanwhile, we tokenize the sentence query into $L$ word tokens and represent the query with a sequence of word embeddings $S=\{s_1, s_2, \cdots, s_L\}$.
On one hand, for visual representations, we basically consider the video Swin transformer~\cite{liu2021video} and the CLIP model. 
Noticed that the video Swin transformer is pre-trained on ImageNet-22K and Kinetics-600.
Additionally, we also explore more visual backbones including 1) video Swin transformer pre-trained on Something-Something V2; 2) R3M \cite{nair2022r3m} pre-trained on Ego4D.
On the other hand, we use the pre-trained BERT model for text representations.

In the second stage, we further encode video and text representations via intra-modal and inter-modal fusion.
First, we concatenate both Swin and CLIP representations along the channel dimension.
Then, two separate transformer encoders are applied on both visual and text modality for intra-modality fusion.
After that, a cross-modal encoder with multiple transformer layers is stacked to capture the inter-modality information across different modalities.
To enhance the temporal order information, we add the positional embedding before feeding these features into the cross-modal encoder.
Through our video-text encoding architecture, long-term temporal orders of a given video can therefore be built.
We set the layer number of individual encoder and cross encoder as 1 and 5 respectively. 
The hidden size of all the transformer layers is set to 512, and the number of self-attention heads is set to 4.

\subsection{Prediction Manner}
Basically we use the anchor-based prediction manner to detect the timestamp of a query.
To be specific, given the output of $T$ embeddings from Sec~\ref{sec:vl_encode}, we manually pre-define $K$ anchors for each embedding.
Assuming $w_k$ is the window size of the $k$-th anchor, the timestamp of $k$-th anchor for $t$-th embedding in the video can be denoted as $[t-\frac{1}{2}w_k, t+\frac{1}{2}w_k]$.
We clip the anchor so as to ensure each one is within the untrimmed video.
We feed $T$ embeddings into two separate prediction heads to predict 1) coordinate regression of all anchors $b \in [1, T]^{KT \times 2}$ and 2) probabilities of the anchors to be a positive proposal $\tau \in [0,1]^{KT}$ respectively. 
Each prediction head consists of a 2-layer MLPs.

Besides the anchor-based approach, we also explore other prediction manners including 1) predicting a start/end timestamps for each sampled frame, which is the same to 2D-TAN (called as TAN-based); 2) directly predicting a proposal for each frame without manual anchors (called as no-manual anchor).
More detailed comparison will be presented in our revised version.

\subsection{Training and Inference}
To train our model, firstly we compute the Intersection over Union score $o$ between each anchor ($\hat{t}^s$, $\hat{t}^e$) and the ground truth timestamp ($t^s$, $t^e$). 
The anchor will be considered as positive only when the score is larger than a threshold.
Then an alignment loss is adopted to align the predicted confidence scores with the IoU score:

\begin{equation}
    \mathcal{L}_{align}=-\frac{1}{KT}\sum_{i=1}^{KT} o_i\log(s_i) +(1-o_i)\log(1-s_i)
\end{equation}

where $o_i$ and $s_i$ are the IoU score label and prediction confidence score of the $i$-th proposal.
We also adopt a temporal boundary loss to promote the precise location of start and end points, which can be denoted as:

\begin{equation}
    \mathcal{L}_{box} = \frac{1}{N_{pos}}\sum_i \mathcal{L}_{L1}(\hat{t}^s_i, t^s_i) + \mathcal{L}_{L1}(\hat{t}^e_i, t^e_i)
\end{equation}

Noticed that $\mathcal{L}_{box}$ is only computed on positive proposals and $N_{pos}$ is the number of positive proposals. $\mathcal{L}_{L1}$ denotes the smooth L1 loss. 
We adopt a hyper-parameter $\mu$ to control the ratio between alignment loss and boundary loss:

\begin{equation}
    \mathcal{L} = \mathcal{L}_{align} + \mu \mathcal{L}_{box}
\end{equation}

We train our model on Ego4D training set with 100 epochs and the batch size is set as 32.
During training, We use Adam optimizer with initialized learning rate 2e-4 and inverse-square-root scheduler.
During inference, given a video and a sentence query, our model samples $T$ frames and predict $KT$ proposals. 
$T$ is set as 600 in our best results.
We select top-5 proposals according to the confidence score as our final results.
\section{Experiments}
We adopt ``R@n, IoU@m'' as the metrics, which is defined as the percentage of at least one of top-$n$ segments having
larger IoU than $m$ with the ground-truth timestamp.
Our final experimental results on the Ego4D test set is demonstrated in Table~\ref{tab:test}.
In the following section, we provide ablation study to show more details of our model.

\begin{table}[h]
\caption{Experimental results on Ego4D test set.}
\label{tab:test}
\centering
\small
\begin{tabular}{c|c |c c |c c }
\toprule
\multirow{2}{*}[-0.5ex]{Visual} & \multirow{2}{*}[-0.5ex]{Text} &\multicolumn{2}{c|}{{IoU=0.3}} & \multicolumn{2}{c}{{IoU=0.5}} \\ 
\cmidrule{3-6}
& & R@1 & R@5 &R@1 & R@5 \\
\midrule

 Swin+CLIP & CLIP & 10.34&18.01 &6.09 & 10.71\\
\bottomrule
\end{tabular}
\end{table}

\subsection{Ablation Study}
\noindent\textbf{How does the anchor scale affect the results?}
We investigate the impact of different anchor scales as shown in Table~\ref{tab:scale}.
We achieve the best results using scales of [0.01, 0.03]. 
In our experiments, predicting more anchors for each temporal index do not bring improvement.
This is mainly due to the excessively long untrimmed video in Ego4D (most videos are longer than 480 seconds). 
As a comparison, the average length of ActivityNet is around 180 seconds, which means we require to sample more than three times of frames on Ego4D therefore overwhelming proposals are inevitable.
To avoid that, we can only reduce the number of anchor scales into 2.

\begin{table}[h]
\caption{Results of different anchor scales on Ego4D val set. Here $r$ represents the proportion of anchor windows.}
\label{tab:scale}
\centering
\small
\begin{tabular}{c|c c |c c }
\toprule
\multirow{2}{*}[-0.5ex]{scale}  &\multicolumn{2}{c|}{{IoU=0.3}} & \multicolumn{2}{c}{{IoU=0.5}} \\ 
\cmidrule{2-5}
& R@1 & R@5 &R@1 & R@5 \\
\midrule
$r=[0.01,0.03,0.09]$ &8.05 &16.65 &3.63 & 8.44 \\
$r=[0.01,0.03]$& 8.33&19.12 &4.39 & 10.69\\
\bottomrule
\end{tabular}
\end{table}

\noindent\textbf{Comparison of different features. }
Table~\ref{tab:feat} demonstrates that our best results are based on video Swin transformer (Kinetics-600) plus CLIP model.
Actually, we have also explored other visual backbones such as R3M and video Swin pre-trained on SSv2.
These backbones are pre-trained on ego-centric datasets like Ego4D (R3M) and SSv2 (video Swin).
To our surprise, previous backbones pre-trained on ego-centric datasets do not contribute to this task in our experiments.

\begin{table}[h]
\caption{Results of different features on the Ego4D val set.}
\label{tab:feat}
\centering
\small
\begin{tabular}{c|c |c |c }
\toprule
\multirow{2}{*}[-0.5ex]{Visual} & \multirow{2}{*}[-0.5ex]{Text} &\multicolumn{1}{c|}{{IoU=0.3}} & \multicolumn{1}{c}{{IoU=0.5}} \\ 
\cmidrule{3-4}
& & R@1 &R@1 \\
\midrule
Slowfast & BERT & 6.61&3.54\\
Swin+CLIP & BERT & 7.20&3.79\\
Swin+CLIP & CLIP & 8.33&4.39\\
\bottomrule
\end{tabular}
\end{table}

\noindent\textbf{Comparison of different sampled video frames. }
Next we explore different sampled video frames in Table~\ref{tab:frame}.
As you can see, Ego4D requires to sample much more frames compared with other traditional benchmark (e.g., 200 frames for ActivityNet and TACoS, 64 frames for Charades-STA).

\begin{table}[h]
\caption{Results of different sampled video frames on the Ego4D val set.}
\label{tab:frame}
\centering
\small
\begin{tabular}{c|c |c }
\toprule
\multirow{2}{*}[-0.5ex]{frames}  &\multicolumn{1}{c|}{{IoU=0.3}} & \multicolumn{1}{c}{{IoU=0.5}} \\ 
\cmidrule{2-3}
& R@1&R@1\\
\midrule
300& 6.38&2.59\\
400& 6.69&2.49\\
600& 8.33&4.39\\
\bottomrule
\end{tabular}
\end{table}

\subsection{Proposal Re-ranking}
We adopt two re-ranking strategies in the challenge.
First, we additionally use CLIP to compute the cosine similarity of representations for each proposal and the sentence query.
Second, we use named entity recognition to extract object entities from the sentence query, then extract object-level features for each frame using MDETR.
Similarly, we use the object feature to compute object-level similarity between each proposal and query.
The CLIP similarity score and the MDETR object-level score will be added to the raw one for the final prediction.

\begin{table}[h]
\caption{Results of proposal re-ranking on the Ego4D val set.}
\label{tab:sta_}
\centering
\small
\begin{tabular}{c|c c |c c }
\toprule
\multirow{2}{*}[-0.5ex]{Method}  &\multicolumn{2}{c|}{{IoU=0.3}} & \multicolumn{2}{c}{{IoU=0.5}} \\ 
\cmidrule{2-5}
& R@1 & R@5 &R@1 & R@5 \\
\midrule
w/o rerank& 8.33& 19.12 & 4.39 & 10.69\\
w/  rerank& 8.85 &20.72 & 4.59 & 11.18\\
\bottomrule
\end{tabular}
\end{table}
\section{Conclusion}
In this work we give a brief introduction of our method for Ego4D natural language query challenge.
We notice that NLQ task on Ego4D faces two major challenges: 1) excessively long untrimmed video vs. short query duration; 2) requirement of long-term temporal orders in the video.
More details will be provided in our revised version.

{\small
\bibliographystyle{ieee_fullname}
\bibliography{reference}
}

\end{document}